\documentclass[sn-apa,iicol]{sn-jnl}

\usepackage{graphicx}%
\usepackage{multirow}%
\usepackage{amsmath,amssymb,amsfonts}%
\usepackage{amsthm}%
\usepackage{mathrsfs}%
\usepackage[title]{appendix}%
\usepackage{xcolor}%
\usepackage{textcomp}%
\usepackage{manyfoot}%
\usepackage{booktabs}%
\usepackage{algpseudocode}%
\usepackage{listings}%
\usepackage{footnote} 

\usepackage{subcaption}      
\usepackage{pifont} 

\usepackage[ruled,vlined]{algorithm2e}  
\usepackage[ruled,vlined]{algorithm2e}
\definecolor{commentcolor}{RGB}{110,154,155}   
\newcommand{\PyComment}[1]{\ttfamily\textcolor{commentcolor}{\# #1}}  
\newcommand{\PyCode}[1]{\ttfamily\textcolor{black}{#1}} %

\begin{document}
 
\title[Open-Vocabulary Object Detection via Neighboring Region Attention Alignment]{Open-Vocabulary Object Detection via Neighboring Region Attention Alignment}

  
\author[1]{\fnm{Sunyuan} \sur{Qiang}}\email{3220004460@student.must.edu.mo}   
 
\author[2]{\fnm{Xianfei} \sur{Li}}\email{felix.li@cowarobot.com} 

\author*[1]{\fnm{Yanyan} \sur{Liang}}\email{yyliang@must.edu.mo}

\author[3]{\fnm{Wenlong}  \sur{Liao}}\email{igoliao@sjtu.edu.cn} 

\author[2]{\fnm{Tao}  \sur{He}}\email{Tommie.he@cowarobot.com}  
  
\author*[2]{\fnm{Pai}   \sur{Peng}}\email{pengpai\_sh@163.com}

\affil[1]{ \orgname{Macau University of Science and Technology}, \orgaddress{\city{Taipa}, \country{Macau SAR}}}

\affil[2]{\orgname{Cowarobot}, \orgaddress{\city{Shanghai}, \country{China}}}  

\affil[3]{\orgname{Shanghai Jiao Tong University}, \orgaddress{\city{Shanghai},  \country{China}}}


\abstract{The nature of diversity in real-world environments necessitates neural network models to expand from closed category settings to accommodate novel emerging categories. In this paper, we study the open-vocabulary object detection (OVD), which facilitates the detection of novel object classes under the supervision of only base annotations and open-vocabulary knowledge. However, we find that the inadequacy of neighboring relationships between regions during the alignment process inevitably constrains the performance on recent distillation-based OVD strategies. To this end, we propose Neighboring Region Attention Alignment (NRAA), which performs alignment within the attention mechanism of a set of neighboring regions to boost the open-vocabulary inference. Specifically, for a given proposal region, we randomly explore the neighboring boxes and conduct our proposed neighboring region attention (NRA) mechanism to extract relationship information. Then, this interaction information is seamlessly provided into the distillation procedure to assist the alignment between the detector and the pre-trained vision-language models (VLMs). Extensive experiments validate that our proposed model exhibits superior performance on open-vocabulary benchmarks. 
}

\keywords{Open-vocabulary object detection, vision-language models, knowledge distillation, attention} 
  

  
\maketitle  
 

\section{Introduction}\label{sec:intro}

\begin{figure*}[!htb] 
   \centering 
  \begin{subfigure}[t]{0.2\linewidth}      
  \includegraphics[width=0.65\linewidth]   {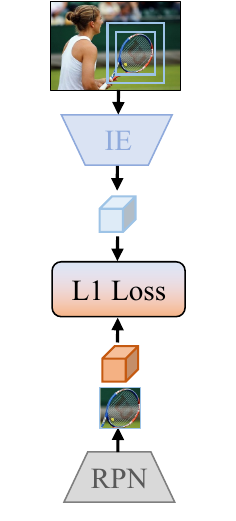}  
  \caption{ViLD. $\qquad$ $\,$ }        
  \label{figure_comparison_vild}
  \end{subfigure}
  \, 
  \begin{subfigure}[t]{0.2\linewidth}  
    \includegraphics[width=0.65\linewidth]    {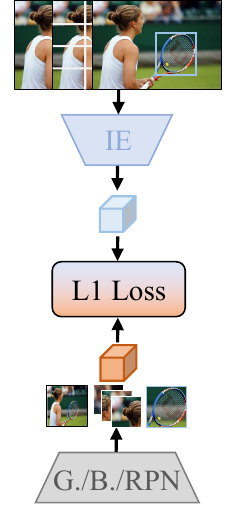}  
  \caption{OADP. $\qquad$ $\quad$ }     
  \label{figure_comparison_oadp}
  \end{subfigure}
  \,  
  \begin{subfigure}[t]{0.2\linewidth}  
    \includegraphics[width=0.65\linewidth]  {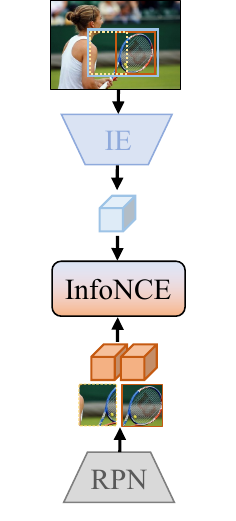}
  \caption{BARON. $\qquad$ $\quad$}        
  \label{figure_comparison_baron}
  \end{subfigure}
  \begin{subfigure}[t]{0.2\linewidth}  
  \includegraphics[width=0.65\linewidth]         {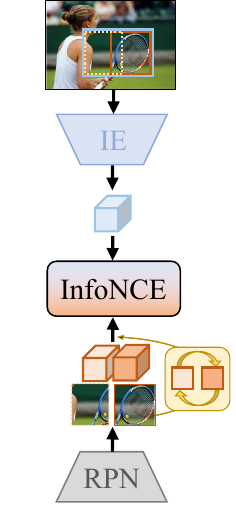}    
  \caption{NRAA (Ours). $\quad$}  
  \label{figure_comparison_ours}  
  \end{subfigure}  
  \caption{The comparison between recent distillation (alignment) based OVD methods and our proposed Neighboring Region Attention Alignment (NRAA). IE and RPN denote the pretrained vision-language image encoder from CLIP and region proposal network. (G.) and (B.) denote the global pooling and block pooling  operators, respectively. Here, we omit the text encoder used in BARON and our method for better visualization. (a) ViLD~\cite{DBLP:conf/iclr/GuLKC22}. (b) OADP~\cite{DBLP:journals/corr/abs-2303-05892}. (c) BARON~\cite{wu2023baron}. (d) NRAA (Ours).  }   
  \label{figure_comparison}  
\end{figure*}

\newcommand\blfootnote[1]{%
  \begingroup
  \renewcommand\thefootnote{}\footnote{#1}%
  \addtocounter{footnote}{-1}%
  \endgroup
} 
\blfootnote{Work done when S. Qiang interned at Cowarobot.}

The inherent intricacies of diversity within real-world environments impose the necessity on deep neural networks (DNNs) to transcend their pre-defined, closed-category settings to adapt to the emergence of novel categories.   
In this work, we embark on an investigation of open-vocabulary object detection (OVD)~\cite{DBLP:conf/cvpr/ZareianRHC21,DBLP:conf/iclr/GuLKC22} that seeks to address the above challenge for the detection of previously unseen object classes using open-vocabulary knowledge.

Conventional object detectors~\cite{DBLP:conf/nips/RenHGS15,DBLP:conf/cvpr/RedmonDGF16,DBLP:conf/eccv/CarionMSUKZ20} are inherently constrained to generalizing recognition exclusively within a predetermined set of object classes, thereby lacking the capability to detect novel unseen categories, which imposes substantial limitations on their utility across diverse applications. In light of this, zero-shot object detection (ZSD)~\cite{DBLP:conf/eccv/BansalSSCD18,DBLP:conf/aaai/RahmanKB20} emerged as a learning framework to detect previously unseen categories. Given the absence of supervised annotations about novel classes, the performance of ZSD has been significantly constrained. 
Finally, owing to the advancements in vision-language models (VLMs)~\cite{DBLP:conf/icml/RadfordKHRGASAM21,DBLP:conf/icml/JiaYXCPPLSLD21}, OVD~\cite{DBLP:conf/cvpr/ZareianRHC21,DBLP:conf/iclr/GuLKC22} leverages additional open-vocabulary sources of supervised information, such as text-image datasets~\cite{DBLP:journals/corr/ChenFLVGDZ15} and pretrained VLMs~\cite{DBLP:conf/icml/RadfordKHRGASAM21}, to improve its ability to detect novel classes. This innovative learning paradigm has garnered considerable attention in recent research community~\cite{DBLP:conf/cvpr/ZareianRHC21,DBLP:conf/iclr/GuLKC22,DBLP:conf/cvpr/DuWZSGL22,kuo2023fvlmopenvocabulary,wu2023baron,DBLP:journals/corr/abs-2303-05892,DBLP:journals/corr/abs-2303-13076}.

\begin{figure}
  \centering    
    
  \begin{subfigure}[t]  {0.49\linewidth}      \includegraphics[width=1.0\linewidth] {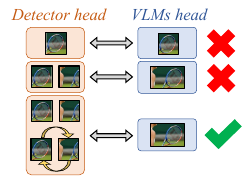}  
  \caption{ }  
  \label{figure_comparison_inconsistency} 
  \end{subfigure}  
  \begin{subfigure}[t]{0.49\linewidth}  
  \includegraphics[width=0.98\linewidth]                 {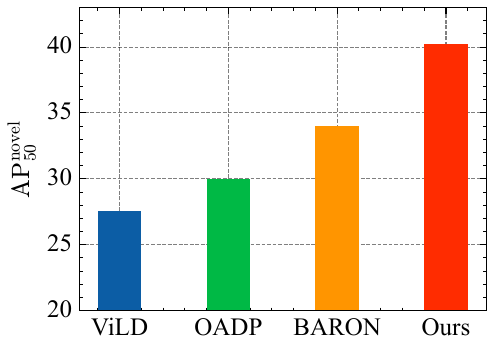}   
  \caption{ }  
  \label{figure_comparison_performance}  
  \end{subfigure}  
  \caption{(a) From top to bottom, the increasing content in VLMs (\textit{e.g.}, expanded inputs) has unearthed more knowledge from the VLMs, and it should be consistent with the yet-to-be-distilled information from the detector during alignment process. (b) Performance comparison of $\text{AP}_{50}^\text{novel}$ metric on OV-COCO benchmarks. }    
  \label{figure_comparison_2}    
\end{figure}

Currently, a typical strategy in OVD is directly to distill (align)~\cite{DBLP:conf/iclr/GuLKC22,DBLP:journals/corr/abs-2303-05892,wu2023baron} open-vocabulary knowledge from VLMs (\textit{e.g.},  CLIP~\cite{DBLP:conf/icml/RadfordKHRGASAM21}) into detector, without supplementary datasets or priors on novel classes, to enhance the detection of novel classes. To be more precise, as shown in Figures~\ref{figure_comparison_vild},~\ref{figure_comparison_oadp}, and~\ref{figure_comparison_baron}, all of these methods are oriented towards maximizing information extraction from the pretrained frozen image encoder (IE) to acquire a broader understanding of CLIP's zero-shot capabilities, such as $1.5\times$ expanded inputs  (ViLD~\cite{DBLP:conf/iclr/GuLKC22}), multi-scale inputs (OADP~\cite{DBLP:journals/corr/abs-2303-05892}), and neighboring expanded inputs (BARON~\cite{wu2023baron}).  
However, \textit{\textbf{these techniques have neglected the inadequacy in the inter-regional interactions information corresponding to the VLM inputs.}} 
Concretely, as shown in Figure~\ref{figure_comparison_inconsistency}, from top to bottom, the increasing content of inputs (\textit{e.g.}, expanded inputs) in VLMs should be consistent with the yet-to-be-distilled information from the detector, and the crux lies in the neglect of insufficient relational information between the regions.

In contrast, our proposed neighboring region attention alignment (NRAA) model in Figure~\ref{figure_comparison_ours}, incorporates an attention mechanism for feature interaction within neighboring regions for distillation alignment. Specifically, given a proposal box from region proposal network (RPN), we firstly draw inspiration from the expanded boxes~\cite{DBLP:conf/iclr/GuLKC22,wu2023baron} to randomly explore neighboring bounding box, aiming to encompass a broader range of visual concept. Then, the neighboring region attention (NRA) mechanism is employed to facilitate interaction between the randomly selected neighboring region features and the proposal box features. Such design endows the model with not only the intrinsic visual concepts of individual regions but also their interconnections for alignment. Finally, we feed these features into the distillation procedure, aligning the detector with the pretrained VLMs
through the infoNCE contrastive loss, demonstrating the exceptional performance as shown in Figure~\ref{figure_comparison_performance}.

Extensive experimental results validate the superior performance of our proposed model. Particularly, using R50 backbone, we achieve a $40.2$ $\text{AP}_{50}^\text{novel}$ and a $21.3$ mask $\text{AP}_r$ on OV-COCO and OV-LVIS, respectively. Furthermore, the results of the ablation studies, as shown in  Tables~\ref{table_ablation_all}  and~\ref{table_ablation_study_main}, distinctly reveal a significant performance improvement, 
with an increase from $25.9$ to $40.2$ ($+14.3$) over the baseline model, 
thus robustly validating its effectiveness.

\section{Related Work}
\label{section_related_work}

\textbf{Open-vocabulary object detection.} Object detection has been a pivotal task in computer vision, traditional approaches~\cite{DBLP:conf/nips/RenHGS15,DBLP:conf/iclr/ZhuSLLWD21,DBLP:conf/iccv/HeGDG17,DBLP:conf/eccv/CarionMSUKZ20} are constrained by the fixed set of object categories, limiting adaptability to previously unseen objects. In response to this, zero-short object detection (ZSD)~\cite{DBLP:conf/eccv/BansalSSCD18,DBLP:conf/aaai/RahmanKB20,DBLP:conf/cvpr/HuangHCZ22,Caixia2022Semantics,DBLP:conf/bmvc/DemirelCI18} firstly emerged as a learning paradigm for detecting novel classes by training only on a set of base classes. Subsequently, 
open-vocabulary detection (OVD)~\cite{DBLP:conf/cvpr/ZareianRHC21,DBLP:conf/iclr/GuLKC22,DBLP:conf/cvpr/DuWZSGL22,kuo2023fvlmopenvocabulary,wu2023baron,DBLP:journals/corr/abs-2303-05892,DBLP:journals/corr/abs-2303-13076} further extended ZSD with significant performance gains under the unified joint image-text vocabulary knowledge (\textit{e.g.}, image-caption dataset~\cite{DBLP:journals/corr/ChenFLVGDZ15}, vision-language models, VLMs~\cite{DBLP:conf/icml/RadfordKHRGASAM21}).

Generally, recent OVD methods can be mainly divided into four themes: (\romannumeral1) \textit{Region-aware training} strategies mainly introduce grounding or contrastive losses to align region and text based on image-text pairs~\cite{DBLP:conf/cvpr/ZareianRHC21,DBLP:conf/cvpr/LiZZYLZWYZHCG22,DBLP:conf/cvpr/KimAK23,DBLP:conf/nips/YaoHWLX0LXX22,DBLP:conf/iclr/LinSJ0QHY023,DBLP:conf/icml/KaulXZ23,DBLP:journals/corr/abs-2309-00775,minderer2023scaling,jin2024llms}. (\romannumeral2) \textit{Pseudo-labeling} methods usually adopt pretrained VLMs~\cite{DBLP:conf/cvpr/ZhongYZLCLZDYLG22,DBLP:conf/eccv/GaoXN0XLX22,DBLP:conf/eccv/ZhaoZSZKSCM22}  or additional classification dataset~\cite{DBLP:conf/eccv/ZhouGJKM22,DBLP:conf/nips/Rasheed0K0K22} to synthesize pseudo region-level labels of novel classes for detection model training. (\romannumeral3) \textit{Knowledge distillation-based} methods directly distill (align) knowledge from VLMs into detector to enable open-vocabulary inference~\cite{DBLP:conf/iclr/GuLKC22,DBLP:conf/cvpr/MaLGLCWZH22,DBLP:conf/cvpr/DuWZSGL22,DBLP:journals/corr/abs-2303-05892,wu2023baron,DBLP:conf/eccv/ZangLZHL22,DBLP:journals/corr/abs-2309-01151,li2023distillingiccv}. (\romannumeral4) \textit{Transfer learning-based} approaches primarily employ the VLMs image encoder as a feature extractor to acquire generalized representations~\cite{kuo2023fvlmopenvocabulary,DBLP:conf/eccv/MindererGSNWDMA22,DBLP:journals/corr/abs-2303-13076,wu2024clipself}.

Among these methods, distillation-based and transfer learning-based methods have garnered significant attention owing to the impressive performance and the capacity to directly harness VLMs without the necessity of supplementary data. In this paper, our proposed NRAA model also shares the same merit of distillation strategy in OVD. Furthermore, we introduce the interaction between neighboring regions of the object, which has been underexplored in current research as shown in Figure~\ref{figure_comparison}, to further enhance the alignment between the VLMs and the detector.

\noindent\textbf{Vision-language models (VLMs) and distillation.} The VLMs~\cite{DBLP:conf/icml/RadfordKHRGASAM21,DBLP:conf/icml/JiaYXCPPLSLD21} align 
visual and language modalities via pretraining with a wealth of image-text pairs, thereby enabling the capability of zero-shot inference in various downstream visual recognition tasks. The first pioneering approach, CLIP~\cite{DBLP:conf/icml/RadfordKHRGASAM21}, leverages contrastive loss to embed textual descriptions and image content into a shared representation space. ALIGN~\cite{DBLP:conf/icml/JiaYXCPPLSLD21} subsequently  scale up data and further enhance performance. Then, numerous techniques have been developed to adapt to downstream task scenarios, where prompt tuning~\cite{DBLP:conf/cvpr/ZhouYL022,DBLP:conf/cvpr/KhattakR0KK23},  adapter~\cite{DBLP:journals/corr/abs-2110-04544,DBLP:conf/eccv/ZhangZFGLDQL22}, and knowledge distillation~\cite{DBLP:conf/cvpr/DongBZZCY00YC0Y23,DBLP:confs/iccv/abs-2307-03135,DBLP:conf/iclr/GuLKC22,DBLP:conf/cvpr/RadenovicDKMVPW23} are three major directions. In this point, open-vocabulary object detection (OVD) can also be considered as a downstream task of VLMs.

Moreover, apart from distillation techniques specifically tailored for VLMs scenarios, there has been significant scholarly attention directed toward more generalized distillation methodologies~\cite{DBLP:journals/corr/HintonVD15,DBLP:conf/iclr/TianKI20,DBLP:conf/cvpr/ZhaoCSQL22,DBLP:journals/corr/abs-2305-15781}. Notably, RKD~\cite{DBLP:conf/cvpr/ParkKLC19} and CC~\cite{DBLP:conf/iccv/PengJLZWLZ019} also devise loss objective functions to introduce the relationships and correlation information for knowledge transfer. Different from all aforementioned methods, our model utilizes the proposed attention mechanism to facilitate relationship interaction between neighboring regions without manually crafted objective functions.

\noindent\textbf{Attention Mechanisms.} Currently, the attention mechanism~\cite{DBLP:conf/nips/VaswaniSPUJGKP17} has demonstrated robust performance across various visual and natural language tasks~\cite{DBLP:conf/iclr/DosovitskiyB0WZ21,DBLP:conf/eccv/CarionMSUKZ20}, further giving rise to numerous modules specially designed for different tasks~\cite{DBLP:conf/iccv/LiuL00W0LG21,DBLP:conf/iccv/0001XMLYMF21,DBLP:conf/iclr/ZhuSLLWD21,DBLP:conf/cvpr/000100LS23}. In object detection, a prevalent class of detectors that harness attention mechanisms is DETR-style detectors~\cite{DBLP:conf/eccv/CarionMSUKZ20},where various attention blocks are typically designed in decoder stage to improve  model performance~\cite{DBLP:conf/iccv/DaiCYZY021,DBLP:conf/iccv/MengCFZLYS021,DBLP:conf/iclr/LiuLZYQSZZ22}. On the other hand, Relation Network~\cite{DBLP:conf/cvpr/HuGZDW18} proposes to model the relationship between object boxes using the object relation module. Human-object interaction~\cite{DBLP:conf/cvpr/GkioxariGDH18} explicitly models the relationships between human and objects using relational verbs.   

On the contrary, our NRAA model aims to explore the attention relationships between neighboring regions of proposal objects for further alignment with VLMs in OVD scenario.

\section{Method}
\label{section_method}

\begin{figure*} [!htb]
  \centering
 \includegraphics[width=0.98\textwidth]   {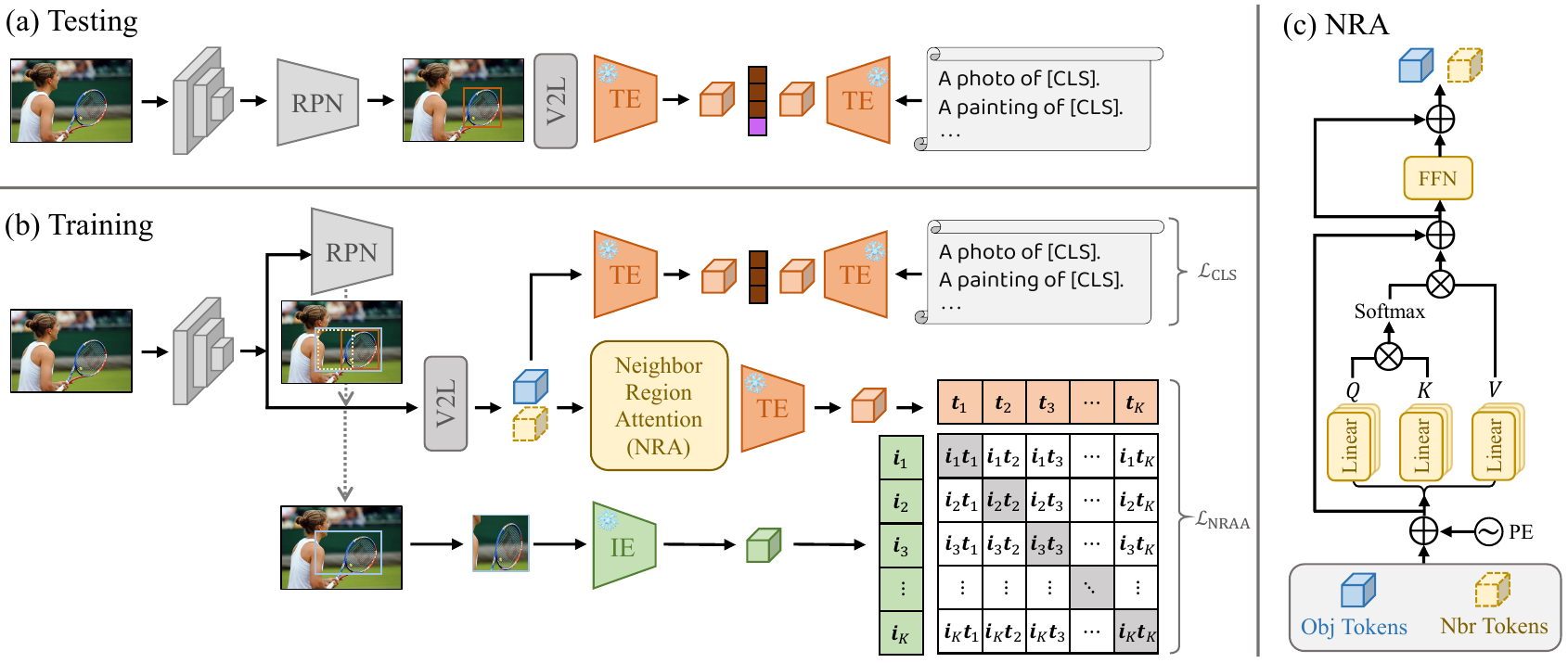}
 \caption{Overview architecture of our proposed NRAA model. (a) Testing stage: NRAA is built upon Faster R-CNN, achieving the OVD classification within the multi-modal text representation space. (b) Training stage: NRAA model introduces an attention mechanism to facilitate interaction among a set of region features for alignment. Here, we omit the basic detection losses for better visualization. (c) Our neighboring region attention (NRA) module. 
 }    
\label{figure_overview}
\end{figure*}

\subsection{Preliminaries} 

\noindent\textbf{Notations.} 
Formally, given an object detection dataset with total categories $C$, we segregate them into base classes $C_B$ and novel classes $C_N$, where $C_B \cap C_N = \emptyset$. The detector model is required to be trained only on base class annotations $C_B$ while achieving the capacity for generalization across the entire categories $C$. 
In addition, a pre-trained vision-language model, \textit{e.g.}, CLIP~\cite{DBLP:conf/icml/RadfordKHRGASAM21}, is also employed, which includes both an image encoder (IE) $\mathcal{I}(\cdot)$ and a text encoder (TE) $\mathcal{T}(\cdot)$.

Our detector model is developed upon the two-stage detector~\cite{DBLP:conf/nips/RenHGS15}. Given an image $\mathbf{x}$ with a proposal region $r$ from region proposal network (RPN), we denote the Region of Interest (RoI) features as 
\begin{equation}
 \mathbf{f}_r = \mathcal{R}(\mathcal{F}(\mathbf{x}), r) ,  
\label{equation_roi_feature}
\end{equation}
where $\mathcal{F}(\cdot)$ and $\mathcal{R}(\cdot)$ denote the backbone model and the RoI feature extractor, respectively.

\noindent\textbf{Motivation.} As stated in Section~\ref{sec:intro}, Figures~\ref{figure_comparison} and \ref{figure_comparison_2}, recent distillation-based OVD methods predominantly focus on expanding the input content into VLMs to optimize the required distilled knowledge, overlooking the inadequacy of inter-regional interactions in the detector. This insight motivates us to incorporate the interrelations between regions into the 
distillation alignment process. Consequently, this further reveals the key challenges addressed in this paper. 
\textbf{(\romannumeral1)} \textit{How to learn the interrelationships among regions and further integrate them into distillation?} \textbf{(\romannumeral2)} \textit{Whether the empirical results confirm the performance in OVD?} \textbf{(\romannumeral3)} \textit{Does the model facilitate information interaction among regions?} We address the aforementioned questions in Section~\ref{section_nraa_model}, Section~\ref{section_ablation_study}, and Figure~\ref{figure_attn_weight_vis}, respectively.

\subsection{NRAA Model}  
\label{section_nraa_model}

The overview architecture of our proposed NRAA model is illustrated in Figure~\ref{figure_overview}. Our model is founded upon the Faster R-CNN~\cite{DBLP:conf/nips/RenHGS15}, extending its capabilities to open-vocabulary inference for novel classes. The pivotal constituents encompass a text classification head, an attention module, and the alignment process.     

\noindent\textbf{Text classification head.} The numerical one-hot classification head is usually substituted with text embeddings $\{\hat{\mathbf{t}}_c\}_{c=1}^{|C|}$ for open-vocabulary inference, which are obtained by providing category names with ensemble of various prompts $w_c$~\cite{DBLP:conf/iclr/GuLKC22} to text encoder $\hat{\mathbf{t}}_c = \mathcal{T} (w_c)$, \textit{e.g.}, $\mathtt{'a \,\, photo \,\, of \,\, [CLS].'}$. We further employ the pseudo words strategy~\cite{wu2023baron}, where RoI features in Equation~\ref{equation_roi_feature} are firstly fed into a linear vision-to-language (V2L) layer for the generation of pseudo words, subsequently subjected to classification via a text encoder. Formally, given a RoI feature $\mathbf{f}_r \in \mathbb{R}^{d_\text{roi}}$ from Equation~\ref{equation_roi_feature}, 
a linear V2L layer maps features into a sequence of tokens with a length of $l$, $\mathbf{f}_r \in \mathbb{R}^{d_\text{roi}} \rightarrow \mathbf{w}_r \in \mathbb{R}^{l \times d_\text{word}}$. In other words, the RoI feature of a object region is represented by a sequence of tokens.    
\begin{equation} 
     \mathbf{w}_r   = \text{Layer}_\text{V2L}( \mathbf{f}_r ) , 
\label{equation_v2l_emb}    
\end{equation} 
where $d_\text{roi}$ and $d_\text{word}$ denote the dimensions of the corresponding features. Then, we concatenate a $\mathtt{[START]}$ token and an $\mathtt{[END]}$ token and further feed the results into the text encoder to obtain the multi-modal text embeddings,   
\begin{equation} 
     \mathbf{t}_r   = \mathcal{T} ( [ \mathtt{[START]} \, ; \,   \mathbf{w}_r \, ; \,      \mathtt{[END]} ]   )    ,   
\label{equation_text_emb} 
\end{equation}
where $[\,;]$ denotes the concatenate operation.    
And the probability of class $c$ for region $r$ is defined as 
\begin{equation} 
    p_{c, r} = \frac{\exp (   \langle \hat{\mathbf{t}}_c, \mathbf{t}_r \rangle / \tau) } {\sum_i \exp ( \langle \hat{\mathbf{t}}_i,  \mathbf{t}_r \rangle / \tau) } ,    
\label{equation_softmax_probability}    
\end{equation}   
where $\langle \, , \rangle$ denotes the cosine similarity and $\tau$ denotes the temperature. $\hat{\mathbf{t}}_c$ is the text embedding associated with the class name $c$  described above. The normalized number of categories during training and testing are $C_B$ and $C$, respectively. Finally, classification loss is defined as 
\begin{equation} 
    \mathcal{L}_\text{CLS} = \frac{1}{|R|}  \sum_{r}^{|R|} \sum_{c}^{C_B} -  ( y_r \cdot  \log  p_{c,r}  )   ,  
\end{equation} 
where $|R|$ is number of proposals per image. $y_r$ denotes the one-hot class label of region $r$.

\begin{figure}[!tb]    
  \centering    
 \includegraphics[width=0.45\textwidth]   {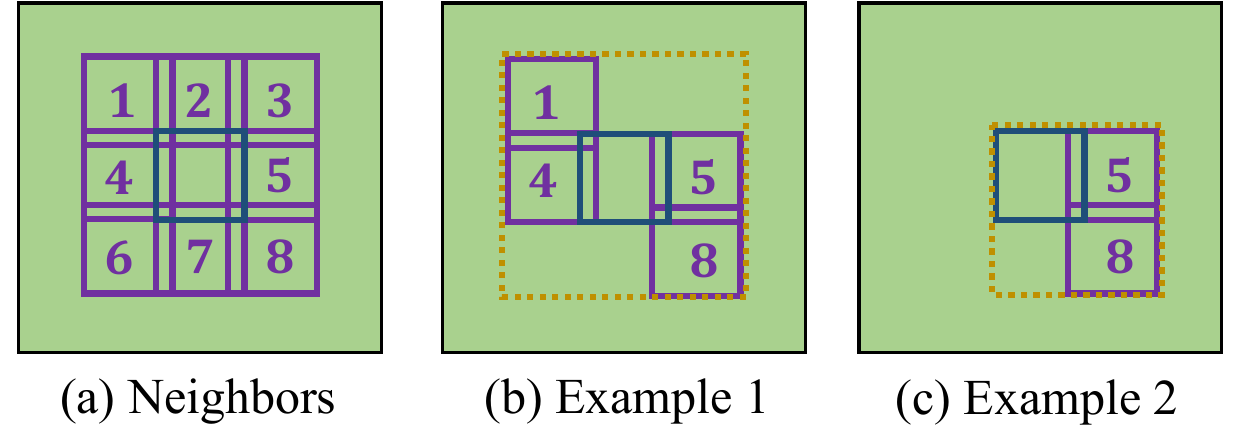}   
 \caption{The visualization results of neighboring regions, where the \textcolor[HTML]{B1D47F}{green} color represents the original image $\mathbf{x}$, the \textcolor[HTML]{2C4D76}{blue} color corresponds to the proposal region $r$, the \textcolor{violet}{violet} color represents neighboring areas labeled with index numbers (1-8), $\{\bar{r}_i\}_{i=1}^8$, and the \textcolor{brown}{brown} dashed lines indicate the outermost expanded region $r_\text{outer}$ fed to the image encoder. Best viewed with color. (a) Neighboring regions of a region proposal. (b) (c) Two sampling  examples. }  
\label{figure_appendix_neighbors}  
\end{figure}

\noindent\textbf{Neighboring region attention (NRA).} As for our proposed neighboring region attention (NRA) module, our goal is to facilitate interaction among the proposal region and its neighboring areas. Inspired by~\cite{DBLP:conf/iclr/GuLKC22,wu2023baron}, for a given proposal $r$ from RPN, we randomly sample regions in the eight surrounding directions $\{\bar{r}_i\}_{i=1}^8$ (top, bottom, left, right, top left, bottom left, top right, bottom right) as shown in Figure~\ref{figure_appendix_neighbors}(a). In addition, the sampled regions must fall within the confines of the image; otherwise, they will be discarded. Thus, we acquire a set of regions $r \cup \{\bar{r}_i\}_{i \in \phi}$, where $r$ is the original proposal box and $\phi$ denotes the set of regional indexes sampled at one time. We can perform multiple random samplings to select neighbors, and the diverse sampling results enhance the richness of contrastive samples in the alignment loss.

Based on the detector model forward process in Equations~\ref{equation_roi_feature} and \ref{equation_v2l_emb}, we obtain a set of tokens corresponding to the sampling regions $\mathbf{w}_r \cup \{ \mathbf{w}_{\bar{r}_i} \}_{i \in \phi}$, where the former are object tokens, while the latter are neighboring tokens. Consequently, this series of tokens can be concatenated to form a sequence of length $l_\phi$, subsequently allowing for interaction through attention mechanisms. Specifically, we have the following expression,  
\begin{equation}
    \begin{matrix} \mathbf{s}_{r,\phi} = \underbrace{  
    [ \mathtt{[START]} \, ; \,\mathbf{w}_r \,;\, \{ \mathbf{w}_{\bar{r}_i} \}_{i \in \phi} \, ; \, \mathtt{[END]} ] }   , \\ 
     \quad \quad \quad 
     {\scriptstyle \text{length} = 1 + l +(l \times |\phi|) + 1 }       
     \end{matrix} 
\label{equation_region_sequence_tokens}  
\end{equation}
where $[\, ; \,]$ is the concatenate operation. $\mathbf{s}_{r,\phi} \in \mathbb{R}^{ (l_\phi+2) \times d_\text{word}}$ denotes our region sequence based on the sampled set $\phi$ and the proposal box $r$. $l_\phi = (|\phi| + 1 ) \times l$ and $|\phi|$ denotes the number of regions in sampled set $\phi$. $l$ is the number of tokens corresponding to one RoI feature as in Equation~\ref{equation_v2l_emb}.

The obtained region sequence $\mathbf{s}_{r,\phi}$ encompass a substantial amount of inherent visual concept, however, they lack sufficient interactive information between region-wise elements. To this end, we introduce the attention mechanism to enable  interaction among tokens. Firstly, we obtain the query $\mathbf{Q}$, key $\mathbf{K}$, and value $\mathbf{V}$ through linear layers $W_Q,W_K,W_V$.    
\begin{equation}
    \begin{aligned}
    \mathbf{Q} = \mathbf{s}_{r,\phi} \cdot W_Q^\top  ,  \, \mathbf{K} = \mathbf{s}_{r,\phi}  \cdot W_K^\top  ,  \,  \mathbf{V} =  \mathbf{s}_{r,\phi} \cdot W_V^\top  .  
    \end{aligned} 
\label{equation_qkv}
\end{equation} 
Then, the scaled dot-product attention operation~\cite{DBLP:conf/nips/VaswaniSPUJGKP17} is preformed as 
\begin{equation}
    {\text{Attn}} (\mathbf{Q},\mathbf{K},\mathbf{V}) = \text{softmax}\left(\frac{\mathbf{Q} \mathbf{K}^\top} {\sqrt{d_\text{word}}}\right) \mathbf{V}.
\label{equation_attention}
\end{equation}  
And the multi-head strategy~\cite{DBLP:conf/nips/VaswaniSPUJGKP17} is further utilized to enhance the capabilities of the model.    

Next, the feed-forward network (FFN) incorporates two linear layers and a GELUs activation function~\cite{DBLP:journals/corr/HendrycksG16} for forward propagation of attention output tokens. Furthermore, positional embeddings and residual shortcuts are also added in our NRA module, as shown in Figure~\ref{figure_overview}(c).

In brief, our NRA model ingests multiple region feature tokens and constructs a series of network components for interactive transformation, ultimately deriving discriminative features. We summarize Equation~\ref{equation_region_sequence_tokens},~\ref{equation_qkv}, and~\ref{equation_attention}, and the aforementioned components (\textit{e.g.}, FFN, PE, shortcuts.) as follows:   
\begin{equation} 
    \hat{\mathbf{s}}_{r,\phi} = \text{NRA}( \mathbf{s}_{r,\phi}  )  ,
\label{equation_nra_forward}
\end{equation} 
where $\text{NRA}$ denotes the whole neighboring region attention module, as shown in Figure~\ref{figure_overview} (c). The refined feature tokens $\hat{\mathbf{s}}_{r,\phi}$ concurrently provide visual concept and relational information for subsequent distillation with VLMs. In this manner, the yet-to-be-distilled information from both the detector and VLM aligns, as shown in Figure~\ref{figure_comparison_inconsistency}, and such knowledge transfer through distillation facilitates the improvement of OVD performance.

\noindent\textbf{Alignment.} The information provided by the detector is the refined feature tokens $\hat{\mathbf{s}}_{r,\phi}$ as discussed above. We feed it into the frozen text encoder $\mathcal{T}(\cdot)$ to obtain multi-modal text embeddings.  
\begin{equation}
    \mathbf{t}_{r,\phi} = \mathcal{T} ( \hat{\mathbf{s}}_{r,\phi}) . 
\end{equation} 

On the other hand, the input of the frozen image encoder $\mathcal{I}(\cdot)$ is the expanded cropped region according to the sampled regions, 
\begin{equation}
    \mathbf{x}_{r, \phi} = \text{crop} ( \mathbf{x},  r \cup \{\bar{r}_i\}_{i \in \phi})  , 
\end{equation} 
where $\text{crop}(  \cdot , \cdot )$ here denotes the outermost cropping operation based on a set of bounding boxes, as shown in Figures~\ref{figure_appendix_neighbors}(b) and~\ref{figure_appendix_neighbors}(c). The expanded region design shares the same advantages in~\cite{DBLP:conf/iclr/GuLKC22,wu2023baron}, providing peripheral information from images for the distillation process. Similarly, we get the multi-modal image embeddings, 
\begin{equation}
    \mathbf{i}_{r,\phi} = \mathcal{I} ( \mathbf{x}_{r, \phi} ) . 
\end{equation} 
Finally, infoNCE loss is performed for alignment. Note that $r$ and $\phi$ denotes indices of the proposals and sample sets. For simplicity, we directly use $K$ to represent the total number of contrastive samples and omit the batch-wise samples.   
\begin{equation} \mathcal{L}_\text{NRAA} = \frac{1}{2} \sum_{k}^K - ( \log (p_{k,1}) + \log (p_{k,2}) )  ,   
\label{equation_kd_infonce_loss}  
\end{equation}
where $p_{k,1}$ and $p_{k,2}$ are the text-to-image ($\mathbf{t}$-to-$\mathbf{i}$) probability and image-to-text ($\mathbf{i}$-to-$\mathbf{t}$) probability, respectively.   
\begin{equation} 
\begin{aligned}
    p_{k,1} &   = \frac{\exp (   \langle \mathbf{i}_k,  \mathbf{t}_k  \rangle / \tau) } {\sum_i \exp (   \langle \mathbf{i}_i, \mathbf{t}_k  \rangle / \tau) } , \\   
    p_{k,2} & = \frac{\exp (   \langle \mathbf{i}_k,  \mathbf{t}_k  \rangle / \tau) } {\sum_j \exp (   \langle \mathbf{i}_k, \mathbf{t}_j \rangle / \tau) } .   
\end{aligned}    
\end{equation}

\begin{table*}[!htb]\small       
\centering  
\caption{Comparison results on OV-COCO benchmark. We employ $\text{AP}_{50}^\text{novel}$ as the quantitative evaluation metric. The comparative methods are categorized into three groups based on the utilization of extra data and the detector framework.  }      
\begin{tabular}{l|cccc|ccc} 
\toprule   
Method & Bbone & Detector & VLMs & Extra Data & $\text{AP}_{50}^\text{novel}$ & \textcolor{gray}{$\text{AP}_{50}^\text{base}$}  & \textcolor{gray}{$\text{AP}_{50}$}  \\ 
\midrule 
OVR-CNN~\cite{DBLP:conf/cvpr/ZareianRHC21} & R50 & F. R-CNN & - & Caption & $22.8$ & \textcolor{gray}{$46.0$} & \textcolor{gray}{$39.9$} \\ 
Detic~\cite{DBLP:conf/eccv/ZhouGJKM22}  & R50 & F. R-CNN & CLIP & Caption & $27.8$ & \textcolor{gray}{-} & \textcolor{gray}{$45.0$}  \\   
CFM-ViT~\cite{DBLP:journals/corr/abs-2309-00775} & ViT-B & M. R-CNN & - & ALIGN & $30.8$ & \textcolor{gray}{-} & \textcolor{gray}{$42.4$}  \\ 
VLDet~\cite{DBLP:conf/iclr/LinSJ0QHY023}  & R50 & F. R-CNN & CLIP & Caption & $32.0$ & \textcolor{gray}{$50.6$} & \textcolor{gray}{$45.8$}  \\   
OC-OVD~\cite{DBLP:conf/nips/Rasheed0K0K22}  & R50 & F. R-CNN & CLIP & Caption & $36.9$ & \textcolor{gray}{$56.6$} & \textcolor{gray}{$51.5$}   \\    
\midrule
DK-DETR~\cite{li2023distillingiccv} & R50 & Def. DETR & CLIP & - & $32.3$ &   \textcolor{gray}{-} & \textcolor{gray}{$61.1$} \\    
CORA~\cite{DBLP:journals/corr/abs-2303-13076}  & R50 & DAB DETR & CLIP & - & $35.1$ & \textcolor{gray}{$35.5$} & \textcolor{gray}{$35.4$} \\   
EdaDet~\cite{DBLP:journals/corr/abs-2309-01151} & R50 & Def. DETR & CLIP & - & $37.8$ & \textcolor{gray}{$57.7$} & \textcolor{gray}{$52.5$}  \\    
\midrule
ViLD~\cite{DBLP:conf/iclr/GuLKC22} & R50 & F. R-CNN & CLIP & - & $27.6$ & \textcolor{gray}{$59.5$} & \textcolor{gray}{$51.3$} \\ 
F-VLM~\cite{kuo2023fvlmopenvocabulary}  & R50 & M. R-CNN & CLIP & - & $28.0$ & \textcolor{gray}{-} & \textcolor{gray}{$39.6$}   \\
OADP~\cite{DBLP:journals/corr/abs-2303-05892}  & R50 & F. R-CNN & CLIP & - & $30.0$ & \textcolor{gray}{$53.3$} & \textcolor{gray}{$47.2$} \\
BARON~\cite{wu2023baron}  & R50 & F. R-CNN & CLIP & - & $34.0$ & \textcolor{gray}{$60.4$} & \textcolor{gray}{$53.5$} \\
\textbf{NRAA (Ours)} & R50 & F. R-CNN & CLIP & - & $\mathbf{40.2}$ & \textcolor{gray}{ $58.6$} &  \textcolor{gray}{$53.8$} \\  
\bottomrule  
\end{tabular} 
\label{table_comparison_ov_coco}   
\end{table*}

\section{Experimental Results}

\subsection{Setup}
\label{section_exp_setup}
   
The experiments are conducted on two widely used object detection datasets, COCO~\cite{DBLP:journals/corr/ChenFLVGDZ15} and LVIS~\cite{DBLP:conf/cvpr/GuptaDG19}. The COCO dataset comprises a total of 80 object categories. In OVD scenario, this dataset is further subdivided into 48 base classes for model training, along with an additional 17 novel classes for open-vocabulary inference, which is commonly referred to as the OV-COCO benchmark. As for the LVIS dataset, it is essentially a long-tail dataset, with categories divided into three distinct levels: frequent, common, and rare. In OVD evaluation protocols, the rare categories (337 classes) are directly excluded and considered as novel classes for open-vocabulary inference. Conversely, the data belonging to the common and frequent categories are retained and treated as base classes for model training. We denote the open-vocabulary LVIS benchmark as OV-LVIS benchmark. 
 
The quantitative metrics primarily revolve around the Average Precision (AP) for novel classes. In the case of OV-COCO, we employ box $\text{AP}_{50}^\text{novel}$ as the main metrics, while in OV-LVIS, we mainly utilize the mask $\text{AP}_r$ for assessment. 
Furthermore, we also provide other detection performance metrics for comprehensive reporting.

\subsection{Implementation Details}

Our model is constructed upon the Faster R-CNN~\cite{DBLP:conf/nips/RenHGS15} architecture and implemented with the PyTorch~\cite{DBLP:conf/nips/PaszkeGMLBCKLGA19}. We elaborate on our implementation details from the following three aspects: 
 
\noindent\textbf{Basic training settings.}  During model training, we employ the SGD optimizer with an initial learning rate of $0.02$, while also  incorporating both the $0.9$ momentum and the $2.5\times10^{-5}$ weight decay. The total batch size is set to $16$ and $32$ for OV-COCO and OV-LVIS, respectively. 
The training schedule is $90k$ iterations, and learning rate is linearly warmed up for the first $1k$ iterations, after which it is divided by $10$ at iteration $60k$ and $80k$.

\noindent\textbf{Model architecture.}  Consistent with previous studies~\cite{DBLP:conf/cvpr/DuWZSGL22}, ResNet-50~\cite{DBLP:conf/cvpr/HeZRS16} and FPN~\cite{DBLP:conf/cvpr/LinDGHHB17} with SoCo~\cite{DBLP:conf/nips/WeiGWHL21} initialization are employed as the backbone network for the detector, and the ViT-B/32 CLIP~\cite{DBLP:conf/icml/RadfordKHRGASAM21} model serves as the OVD distillation alignment teacher.

\noindent\textbf{Alignment settings.} For the infoNCE alignment loss, we maintain two queues throughout training to retain image and text embeddings from previous iterations as negative samples. With respect to the temperature coefficient $\tau$, it predominantly involves three facets: training classification loss, testing classification inference, and alignment loss. We set these coefficients to $\frac{1}{50}$, $\frac{1}{50}$, $\frac{1}{30}$, and $\frac{1}{100}$, $\frac{3}{400}$, $\frac{1}{20}$ on OV-COCO and OV-LVIS benchmarks, respectively. Moreover, the ensemble of manually crafted prompts~\cite{DBLP:conf/iclr/GuLKC22} are utilized in OV-COCO, whereas the learned prompts~\cite{DBLP:conf/cvpr/DuWZSGL22} are employed in OV-LVIS. As for the vision-to-language layer, same as the~\cite{wu2023baron}, the RoI feature $\mathbf{f}_r$ is mapped to word tokens $\mathbf{w}_r$ with lengths $l$ of $6$ and $4$ in OV-COCO and OV-LVIS, respectively,  followed by the dropout to randomly exclude specific tokens with probability $0.5$. In OV-LVIS, we also add additional individual loss and expanded regions for further alignment.

\subsection{Main Results}

We compare our proposed method with various existing models as shown in Tables~\ref{table_comparison_ov_coco},~\ref{table_comparison_ov_lvis} and~\ref{table_transfer_dataset}. The experimental results consistently demonstrate the robust performance enhancements achieved by our NRAA model.

\begin{table*}[!htb]\small
\centering
\caption{Comparison results on OV-LVIS benchmark. We employ both box $\text{AP}_r$ and mask $\text{AP}_r$ as the quantitative evaluation metric. 
The symbol $\dagger$ and $*$ denote the utilization of double heads ensemble and extra data. 
}     
\begin{tabular}{l|c|cccc|cccc} 
\toprule
\multirow{2}{*}{Method} & \multirow{2}{*}{BBone} & \multicolumn{4}{c|}{Object Detection} & \multicolumn{4}{c}{Instance Segmentation} \\   
 & & $\text{AP}_r$ & \textcolor{gray}{$\text{AP}_c$} & \textcolor{gray}{$\text{AP}_f$} & \textcolor{gray}{$\text{AP}$} & $\text{AP}_r$ & \textcolor{gray}{$\text{AP}_c$} & \textcolor{gray}{$\text{AP}_f$} &  \textcolor{gray}{$\text{AP}$} \\  
\midrule   
ViLD~\cite{DBLP:conf/iclr/GuLKC22} & R50 & $16.3$ & 
\textcolor{gray}{$21.2$} & 
\textcolor{gray}{$31.6$} & 
\textcolor{gray}{$24.4$} & $16.1$ & \textcolor{gray}{$20.0$} & \textcolor{gray}{$28.3$} & \textcolor{gray}{$22.5$}  \\ 
ViLD$^\dagger$~\cite{DBLP:conf/iclr/GuLKC22} & R50 & $16.7$ & 
\textcolor{gray}{$26.5$} & 
\textcolor{gray}{$34.2$} & 
\textcolor{gray}{$27.8$} & $16.6$ & \textcolor{gray}{$24.6$} & \textcolor{gray}{$30.3$} & \textcolor{gray}{$25.5$} \\  
BARON~\cite{wu2023baron} & R50 & $17.3$ &   \textcolor{gray}{$25.6$} & \textcolor{gray}{$31.0$} &   \textcolor{gray}{$26.3$} & $18.0$ & \textcolor{gray}{$24.4$} & \textcolor{gray}{$28.9$} & \textcolor{gray}{$25.1$}  \\   
RegionCLIP$^*$~\cite{DBLP:conf/cvpr/ZhongYZLCLZDYLG22} & R50 & - & \textcolor{gray}{-} & \textcolor{gray}{-} &\textcolor{gray}{-} & $17.1$ & \textcolor{gray}{$27.4$} & \textcolor{gray}{$34.0$} & \textcolor{gray}{$28.2$}   \\ 
Detic$^*$~\cite{DBLP:conf/eccv/ZhouGJKM22}  & R50 & - & \textcolor{gray}{-} & \textcolor{gray}{-} & \textcolor{gray}{-} & $17.8$ & \textcolor{gray}{$26.3$} & \textcolor{gray}{$31.6$} & \textcolor{gray}{$26.8$}   \\  
DetPro$^\dagger$~\cite{DBLP:conf/cvpr/DuWZSGL22} & R50 & $20.8$ & \textcolor{gray}{$27.8$} & \textcolor{gray}{$32.4$} & \textcolor{gray}{$28.4$} & $19.8$ & \textcolor{gray}{$25.6$} & \textcolor{gray}{$28.9$} & \textcolor{gray}{$25.9$}  \\ 
OADP$^\dagger$~\cite{DBLP:journals/corr/abs-2303-05892} & R50 & $20.7$ & \textcolor{gray}{-} & \textcolor{gray}{-} &   \textcolor{gray}{-} & $19.9$  & \textcolor{gray}{-} & \textcolor{gray}{-} & \textcolor{gray}{-} \\ 
F-VLM~\cite{kuo2023fvlmopenvocabulary} & R50 & - & \textcolor{gray}{-} & \textcolor{gray}{-} & \textcolor{gray}{-} & $18.6$ & \textcolor{gray}{$24.0$} & \textcolor{gray}{$26.9$} & \textcolor{gray}{$24.2$}  \\   
\textbf{NRAA (Ours)} & R50 & $\mathbf{21.1}$ & \textcolor{gray}{$27.0$} & \textcolor{gray}{$31.7$} & \textcolor{gray}{$27.8$} & $\mathbf{21.3}$ & \textcolor{gray}{$25.1$} & \textcolor{gray}{$28.4$}  & \textcolor{gray}{$25.7$}   \\  
\bottomrule  
\end{tabular}
 \label{table_comparison_ov_lvis}  
\end{table*}

\noindent\textbf{OV-COCO benchmark.} As shown in Table~\ref{table_comparison_ov_coco}, for a fair comparison, the compared methods are classified into three groups based on the incorporation of additional data and the detector framework. The extra datasets typically introduce prior knowledge of novel classes with pseudo labeling, while the latter detection framework primarily adopts the DETR-style architecture. In contrast, our fully distillation based alignment method is built upon the Faster R-CNN~\cite{DBLP:conf/nips/RenHGS15} and requires no additional data. Obviously, using R50 backbone, our model consistently outperforms all previous methods on OV-COCO benchmark,
exhibiting a substantial increase in $\text{AP}_{50}^\text{novel}$ from $34.0$ to $40.2$ ($+6.2$). In extended comparisons with specific pseudo-labeling or DETR based OVD methods, our model still maintains a performance superiority,  surpassing their respective $\text{AP}_{50}^\text{novel}$ of $36.9$ and $37.8$. On the other hand, it is noteworthy that our model also exhibits competitive detection performance on base classes of $58.6$ $\text{AP}_{50}^\text{base}$. These empirical observations not only validate the capability of our model in open-vocabulary inference of novel classes, but also establish the absence of any substantial compromise in the detection performance of the base classes on OV-COCO benchmark.

\begin{table}[!htb]\footnotesize 
\centering  
\caption{Comparison results on transfer dataset benchmark. We evaluate OV-COCO trained model on LVIS. } 
\begin{tabular}{l|ccc}  
\toprule 
\multirow{2}{*}{Method} & \multicolumn{3}{c}{Transfer Dataset} \\ 
 & $\text{AP}$ & $\text{AP}_{50}$ & $\text{AP}_{75}$ \\ 
\midrule 
OVR-CNN~\cite{DBLP:conf/cvpr/ZareianRHC21} & - & $5.2$ & -  \\
OC-OVD~\cite{DBLP:conf/nips/Rasheed0K0K22} & $5.6$ & $8.5$ &  $6.0$ \\
Detic~\cite{DBLP:conf/eccv/ZhouGJKM22} & $5.5$ &  $8.5$ & $5.8$ \\
VLDet~\cite{DBLP:conf/iclr/LinSJ0QHY023} & - & $10.0$ & - \\  
\textbf{NRAA (Ours)} & $8.1$ &  $13.0$ & $8.5$  \\
\bottomrule
\end{tabular} 
\label{table_transfer_dataset}
\end{table}

\noindent\textbf{OV-LVIS benchmark.} For the OV-LVIS, the compared methods mainly utilize the Mask R-CNN architecture, along with some approaches incorporating ensemble heads and extra datasets techniques leveraging novel class priors. As shown in Table~\ref{table_comparison_ov_lvis}, using R50 backbone, we achieve performance with box $\text{AP}_r$ and mask $\text{AP}_r$ of $21.1$ and $21.3$, respectively, surpassing the previous results, which validates the effectiveness of our method. It is noteworthy that our model does not necessitate the utilization of any additional data or the premature leakage of information regarding novel classes.

\noindent\textbf{Transfer dataset.} To further validate the generalization capability of our model, we proceed to evaluate the model trained on OV-COCO directly on the LVIS dataset by replacing the embedding of classifier head. As shown in Table~\ref{table_transfer_dataset}, our model achieves $8.1$ $\text{AP}$, $13.0$ $\text{AP}_{50}$, and $8.5$ $\text{AP}_{75}$, respectively, demonstrating the effectiveness of our model in transfer generalization capability.

\subsection{Ablation Study}
\label{section_ablation_study}

In this subsection, we conduct an exhaustive ablation study to validate the effectiveness of our proposed NRAA model. We begin by presenting the main results and providing a component-wise analysis of the NRA mechanism. Subsequently, we conduct a comprehensive analysis of the alignment architecture employed by the model.

\begin{table*}[!htb]\small   
\centering  
\begin{minipage}[t]{.454\linewidth}       
\caption{Ablation results of neighbors (Eq.~\ref{equation_region_sequence_tokens}) and NRA (Eq.~\ref{equation_nra_forward}) mechanisms on OV-COCO benchmark.   }    
\begin{tabular}{c|cc|ccc}     
\toprule
\# & Nbr. & NRA &  $\text{AP}_{50}^\text{novel}$ & \textcolor{gray} {$\text{AP}_{50}^\text{base}$} & $\Delta$ \\ 
\midrule   
 1 & \multicolumn{2}{c|}{No Distill} & $0.0$ & \textcolor{gray}{$58.2$} & - \\ 
 2 &   - & - & $25.9$ & \textcolor{gray}{$52.3$} & -  \\  
\midrule  
 3 & \ding{51} & & $31.1$ & \textcolor{gray}{$57.1$} & $+5.2$ \\  
 4 & & \ding{51} & $36.2$ & \textcolor{gray}{$55.5$} & $+10.3$  \\ 
 5 & \ding{51} & \ding{51} & $40.2$ & \textcolor{gray}{$58.6$} & $+14.3$ \\ 
\bottomrule 
\end{tabular}   
\label{table_ablation_all}   
\end{minipage}  
\hfill  
\begin{minipage}[t]  {.518\linewidth} 
\caption{Ablation results of our proposed neighboring region attention (NRA) mechanism on OV-COCO benchmark.   } 
\begin{tabular}{c|ccc|ccc}   
\toprule   
\# & N-Attn & FFN & PE & $\text{AP}_{50}^\text{novel}$ & \textcolor{gray}{$\text{AP}_{50}^\text{base}$}  & $\Delta$ \\  
\midrule  
1 & - & - & - & $31.1$ & \textcolor{gray}{$57.1$} & - \\
2 & \ding{51} & & & $38.5$ &  \textcolor{gray}{$57.4$} & $+7.4$  \\    
3 & \ding{51} & \ding{51} & & $39.9$ & \textcolor{gray}{$58.4$} & $+8.8$  \\ 
4 & \ding{51} & \ding{51} & \ding{51}  & $40.2$ & \textcolor{gray}{$58.6$} & $+9.1$ \\   
\bottomrule  
\end{tabular}   
\label{table_ablation_study_main}
\end{minipage}   
\end{table*}

\begin{table*}[!htb]\small    
\centering  
\begin{minipage}[t]{.278\linewidth}      
\caption{Ablation results of the number of proposals (top-$k$) for alignment.  } 
\begin{tabular}{c|cc}    
\toprule 
Top-$k$ & $\text{AP}_{50}^\text{novel}$ & \textcolor{gray}{$\text{AP}_{50}^\text{base}$} \\  
\midrule    
$10$  & $38.1$ & \textcolor{gray}{$54.9$}  \\  
$50$  & $39.0$ & \textcolor{gray}{$57.9$}  \\
$100$ & $39.3$ & \textcolor{gray}{$58.0$}  \\    
$300$ & $40.2$ & \textcolor{gray}{$58.6$}  \\ 
$600$ & $39.0$ & \textcolor{gray}{$58.6$}  \\  
\bottomrule    
\end{tabular} 
\label{table_ablation_study_number_proposals_topk} 
\end{minipage}
\hfill     
\begin{minipage}[t]{.286\linewidth}         
\caption{Ablation results of the number of sampling neighbors. }  
\begin{tabular}{c|cc}  
\toprule  
\# Nbr. & $\text{AP}_{50}^\text{novel}$ & \textcolor{gray}{$\text{AP}_{50}^\text{base}$} \\  
\midrule     
$1$ & $36.6$ & \textcolor{gray}{$55.3$}  \\  
$2$ & $38.5$ & \textcolor{gray}{$57.2$} \\ 
$4$ & $40.2$ & \textcolor{gray}{$58.1$}  \\  
$8$ &  $40.2$ & \textcolor{gray}{$58.6$}  \\    
\bottomrule    
\end{tabular} 
\label{table_ablation_number_of_neighbors}   
\end{minipage}
\hfill  
\begin{minipage}[t]{.27\linewidth}    
\caption{Ablation results of using one single expanded neighbor for alignment. }   
\begin{tabular}{c|cc} 
\toprule  
Coef. & $\text{AP}_{50}^\text{novel}$ & \textcolor{gray}{$\text{AP}_{50}^\text{base}$} \\   
\midrule   
 - & $36.2$ & \textcolor{gray}{$55.5$}  \\  
$0.5$x & $34.6$ & \textcolor{gray}{$56.2$}  \\  
$1.5$x & $36.9$ & \textcolor{gray}{$56.6$}  \\    
$2.0$x & $37.2$ & \textcolor{gray}{$56.0$}  \\  
Ours & $40.2$ & \textcolor{gray}{$58.6$} \\  
\bottomrule    
\end{tabular} 
\label{table_ablation_expanded_box}     
\end{minipage}
\end{table*}

\begin{table*}[!htb]\small    
\centering 
\begin{minipage}[t] {.41\linewidth}     
\caption{Ablation results of using both surrounding and expanded neighbors for alignment. }   
\begin{tabular}{c|cc|cc}  
\toprule  
\# & Surr. & Expand   & $\text{AP}_{50}^\text{novel}$ & \textcolor{gray}{$\text{AP}_{50}^\text{base}$} \\   
\midrule   
 1 & - & - & $36.2$ & \textcolor{gray}{$55.5$}  \\  
 2 & \ding{51} & \ding{51} ($1.5$x) & $39.7$ & \textcolor{gray}{$58.2$}  \\ 
 3 & \ding{51} & \ding{51} ($2.0$x) & $39.0$ & \textcolor{gray}{$57.7$}  \\    
 4 & \ding{51} & - & $40.2$ & \textcolor{gray}{$58.6$} \\  
\bottomrule     
\end{tabular} 
\label{table_ablation_study_both_expand_and_neighbors} 
\end{minipage}
\qquad
\begin{minipage}[t]{.428\linewidth}  
\caption{Ablation results for multi-layer stacked NRA. }
\begin{tabular}{c|cc|cc} 
\toprule  
\# & Layers & Shortcut &  $\text{AP}_{50}^\text{novel}$ & \textcolor{gray}{$\text{AP}_{50}^\text{base}$} \\ 
\midrule     
1 & $1$ & & $40.2$ & \textcolor{gray}{$58.6$} \\
2 & $5$ & & $9.7$ &  \textcolor{gray}{$58.9$} \\ 
3 & $5$ & \ding{51} & $35.7$ & \textcolor{gray}{$58.4$} \\ 
4 & $7$ & \ding{51} & $25.9$  & \textcolor{gray}{$58.5$} \\ 
\bottomrule    
\end{tabular}
\label{table_ablation_study_layers} 
\end{minipage}   
\end{table*}

\noindent\textbf{The NRA mechanism.} Tables~\ref{table_ablation_all} and~\ref{table_ablation_study_main} presents the main ablation results of our proposed neighboring region attention (NRA) mechanism. In Table~\ref{table_ablation_all}, the baseline model follows the idea of distillation alignment through one contrastive loss. Such strategy (\#2) achieves $25.9$ $\text{AP}_{50}^\text{novel}$ and $52.3$ $\text{AP}_{50}^\text{base}$. Simultaneously, models lacking distillation (\#1) prove incapable of predicting novel categories, which validates the performance potential of distillation-based approaches. Introducing neighboring boxes (\#3) for alignment leads to the extraction of more knowledge from pre-trained CLIP model, resulting in $31.1$ $\text{AP}_{50}^\text{novel}$. On the other hand, a noteworthy performance improvement is also observed when employing NRA (\#4) for self-interaction, achieving $36.2$ on $\text{AP}_{50}^\text{novel}$. Finally, when integrating both the neighbors and the NRA (\#5), our model’s performance exhibited a further enhancement, yielding $\text{AP}_{50}^\text{novel}$ and $\text{AP}_{50}^\text{base}$ of $40.2$ and $58.6$, respectively.

As discussed in Section~\ref{section_nraa_model} and Figure~\ref{figure_overview}, our NRA can be partitioned into three components: neighboring self-attention module (N-Attn), feed-forward network (FFN), and positional embeddings (PE). Therefore, we provide detailed analysis of each component in Table~\ref{table_ablation_study_main}. By introducing N-Attn module to the model, the 
$\text{AP}_{50}^\text{novel}$ is significantly improved from $31.1$ to $38.5$ ($+7.4$), indicating its efficacy in boosting model performance on novel classes. Next, when integrating the feed-forward network and positional embeddings, our model's performance exhibited a further enhancement, yielding $\text{AP}_{50}^\text{novel}$ of $39.9$ and $40.2$, respectively. 
In terms of detection performance on the base classes, our model still demonstrates a slight enhancement, specifically elevating from $57.1$ to $58.6$ ($+1.5$) $\text{AP}_{50}^\text{base}$. These empirical observations suggest that our NRA not only contributes to the recognition of novel classes but also does not compromise the performance of base classes.

\begin{figure*}[!tb]  
 \centering 
 \includegraphics[width=0.95\textwidth] {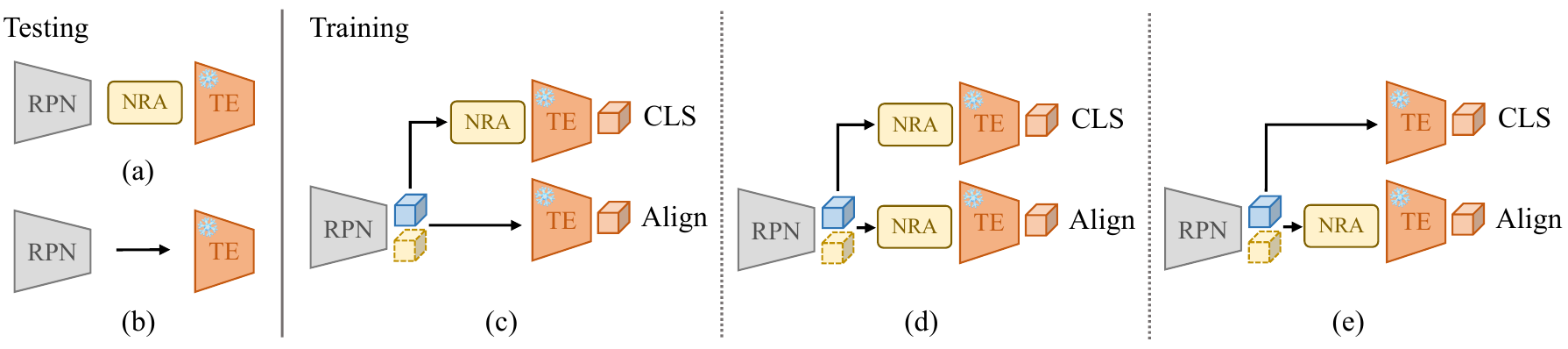}     
 \caption{The ablation architectures of NRA module, serving as an explanatory note for the configuration settings in Table~\ref{table_appendix_ablation_study_nra_pos}. The configuration of the NRA module is categorized into training and testing phases. During the testing stage, only classification needs to be considered,  leading to two scenarios, denoted as (a) and (b), namely, using the NRA module and not using the NRA module. In the training process, both classification and alignment need to be simultaneously considered. Consequently, there are three settings for integrating the NRA module: adding it solely to the classification (c), adding it to both positions (d), and adding it solely to the alignment (e).  }  
\label{figure_appendix_alignment_arch}  
\end{figure*}  
\begin{table*}[!tb]\small  
\centering
\caption{Ablation results for the positions employed by the NRA module. We further design alternative configurations for comprehensive comparisons, and the ablation architectures during training and testing stage using NRA for alignment and classification is illustrated in Figure~\ref{figure_appendix_alignment_arch}.  
}    
\begin{tabular}{c|cc|cc} 
\toprule    
\multirow{2}{*}{\#} & \multicolumn{2}{c|}{Arch in Figure~\ref{figure_appendix_alignment_arch}} & \multirow{2}{*}{$\text{AP}_{50}^\text{novel}$} & \multirow{2}{*}{\textcolor{gray}{$\text{AP}_{50}^\text{base}$}}  \\ 
 & \,\,\, Test \,\, & \,\,  Train \,\,\,  & & \\  
\midrule     
1 & - & - & $31.1$ & \textcolor{gray}{$57.1$} \\  
2 & (a) & (c) & $28.6$ &  \textcolor{gray}{$58.1$} \\ 
3 & (a) & (d) & $39.1$ &  \textcolor{gray}{$58.4$}  \\    
4 & (a) & (e) & $1.3$ & \textcolor{gray}{$2.8$} \\  
5 & (b) & (c) & $17.7$ & \textcolor{gray}{$24.5$} \\  
6 & (b) & (d) & $17.5$ & \textcolor{gray}{$30.9$} \\  
7 & (b) & (e) & $40.2$ &  \textcolor{gray}{$58.6$} \\  
\bottomrule
\end{tabular}   
\label{table_appendix_ablation_study_nra_pos}
\end{table*}

\noindent\textbf{Neighbors strategy.} We extend our ablations on neighbors mechanism, including RPN top-$k$ box filter (Table~\ref{table_ablation_study_number_proposals_topk}), ablation on the number of neighbors (Table~\ref{table_ablation_number_of_neighbors}), and expanded neighbors (Tables~\ref{table_ablation_expanded_box} and~\ref{table_ablation_study_both_expand_and_neighbors}). 
The RPN generates proposal boxes, and top-$k$ boxes are selected based on their scores for subsequent NRA. In Table~\ref{table_ablation_study_number_proposals_topk}, increasing $k$ will bring performance gains. However, a slight performance decline is observed when $k=600$. We posit that the quality of the proposal boxes is equally crucial for NRA, the proposal boxes with lower scores may not contribute significantly. As shown in Table~\ref{table_ablation_number_of_neighbors}, an increased max number of neighbors contribute to a greater infusion of region-based contextual information, and the ablations also observed performance gains. When number is set to $4$ and $8$, the performance remains consistent in $\text{AP}_{50}^\text{novel}$. This suggests that, for contextual information, $4$ neighbors are essentially sufficient. Considering $\text{AP}_{50}^\text{base}$, we opt for $8$ neighbors. 
    
In Table~\ref{table_ablation_expanded_box}, the ablations include both expanded box ($1.5$x and $2.0$x) and reduced box ($0.5$x). Reduced boxes fail to provide effective contextual information, leading to a decline in overall performance. The utilization of expanded boxes yields lower performance in comparison to our model. Similarly, as outlined in Table~\ref{table_ablation_study_both_expand_and_neighbors}, employing both surrounding and expanded neighbors simultaneously also does not exhibit a distinct advantage over utilizing surrounding neighbors independently. We suspect that encompassing a global neighborhood region might lead to insufficient sensitivity to detailed regions and potentially introduce additional noise, thereby contributing to inferior performance compared to our model.

\noindent\textbf{NRA architecture.} Our model demands careful deliberation on how to effectively embed NRA module into the alignment for open-vocabulary inference. In this paper, the NRA module is essentially only employed in the distillation alignment during training. We additionally design configurations for comprehensive comparisons as shown in Table~\ref{table_appendix_ablation_study_nra_pos} and Figure~\ref{figure_appendix_alignment_arch}.

\begin{figure*}[!htb] 
  \centering
 \includegraphics[width=0.98\textwidth]   {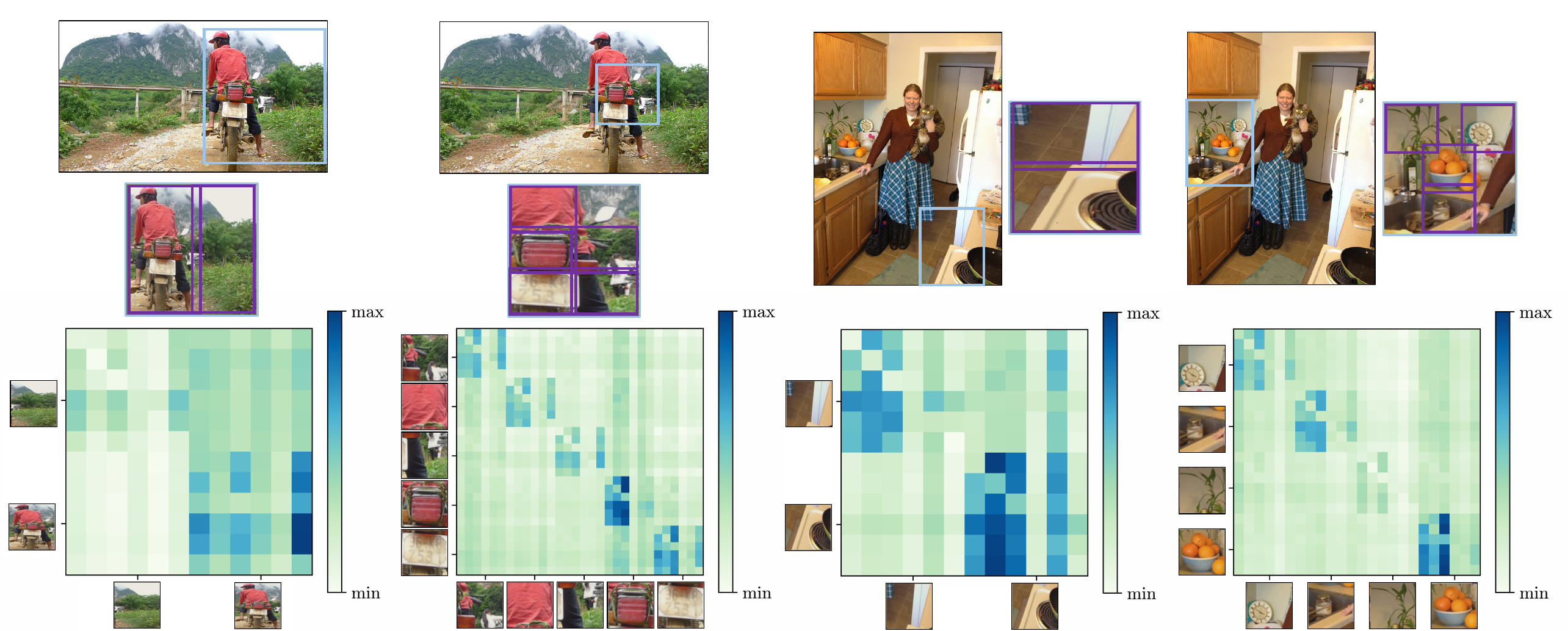}  
 \caption{The heatmap visualization results of attention weights generated by our proposed NRA module. 
  Here, each individual region features are pre-transformed into a sequence of tokens with a length of six, $l = 6$, and we omit the $\mathtt{[START]}$ token and the $\mathtt{[END]}$ token. The \textcolor{cyan}{blue} bounding boxes in the detection images denote the expanded boxes formed by combining all \textcolor{violet}{violet} individual bounding boxes. Best viewed with color. }  
\label{figure_attn_weight_vis}
\end{figure*}

There are a total of $2\times3=6$ combinations of configurations. When employing NRA module in the classification head during concurrent training and testing (\#2), a discernible performance decrease $28.6$ $\text{AP}_{50}^\text{novel}$ is observed compared to the baseline model. Intuitively, the NRA is designed for interaction between regions in alignment head, rather than for discerning visual concepts in classification head. As for (\#3), due to the concurrent utilization of the NRA module in the classification and alignment heads during training, our model effectively enhances performance under such inductive bias with a $\text{AP}_{50}^\text{novel}$ of $39.1$, which validates the effectiveness of NRA module and the seamless integration capability with vision-language alignment. As for (\#4), applying the NRA module learned from the alignment process in classification solely during testing stage results in a significant performance degradation, yielding only a $1.3$ $\text{AP}_{50}^\text{novel}$. If the classification head with NRA module is used in testing, it is essential to train the classification head with NRA in the corresponding training phase. Otherwise, adding NRA module trained only with the alignment head into the testing classification head will evidently result in severe performance degradation. Similarly, for (\#5) and (\#6), the addition of NRA to the classification head during training, followed by its exclusion during testing, resulted in performance degradation due to the inconsistency in the classification head. On the contrary, by excluding the use of the NRA module in classification head and restricting its application solely to alignment head (\#7), we finally achieve  performance of a $40.2$ $\text{AP}_{50}^\text{novel}$. On one hand, this implicitly validates that our model effectively establishes interrelations during the alignment process, promoting the transfer of VLMs knowledge without affecting the inherent visual concepts. Secondly, model requires no additional parameters during inference, highlighting its efficiency.

\noindent\textbf{Stacked NRA modules.} Given that the model does not necessitate the utilization of the NRA module during the inference, we proceeded to stack multiple layers of NRA blocks to systematically explore the ensuing performance variations. 

As shown in Table~\ref{table_ablation_study_layers}, our primary consideration centers around configurations employing 5 and 7 layers. We also design additional residual shortcuts spanning from the initial to the final layer for maintaining the unaltered visual concepts from original inputs. Firstly, we observe that as the number of layers increased, the preservation of original visual knowledge, such as through residual connections, became crucial for maintaining performance. Particularly, we achieve performances of $35.7$ and $9.7$ $\text{AP}_{50}^\text{novel}$, respectively, concerning the inclusion or exclusion of shortcut connections.
Then, the increased interaction of attention blocks, however, tended to further compromise the model's performance. Employing just a single block not only facilitates feature interaction among regions but also preserves the original visual concepts, resulting in the optimal performance of $40.2$ $\text{AP}_{50}^\text{novel}$.

\begin{figure*}
  \centering 
  \begin{subfigure}{0.49\linewidth}
  \includegraphics[width=1.0\linewidth]  {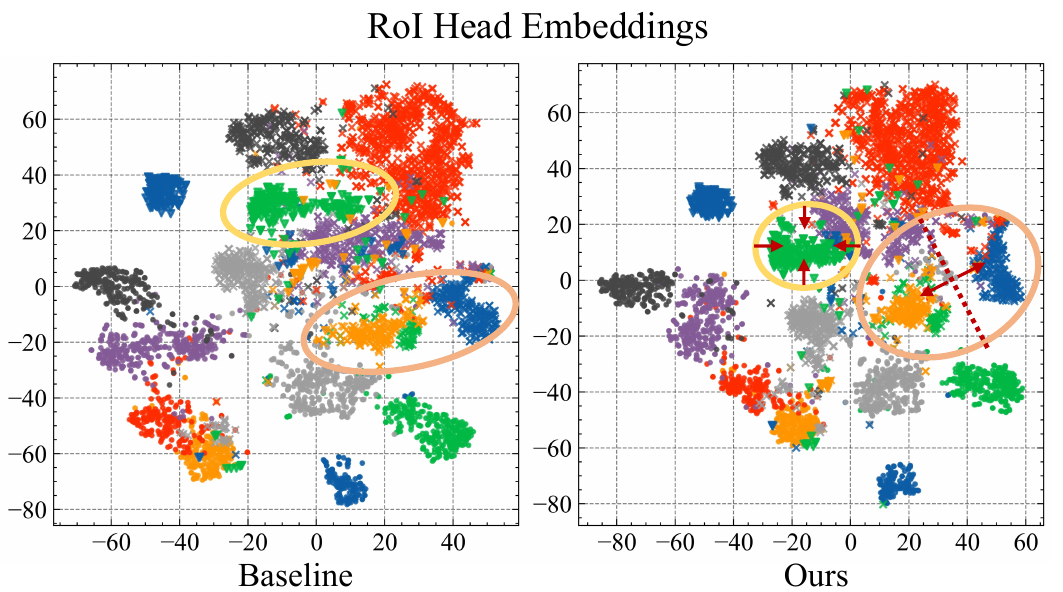}
  \end{subfigure}
  \,   
  \begin{subfigure}{0.49\linewidth}   
  \includegraphics[width=1.0\linewidth]  {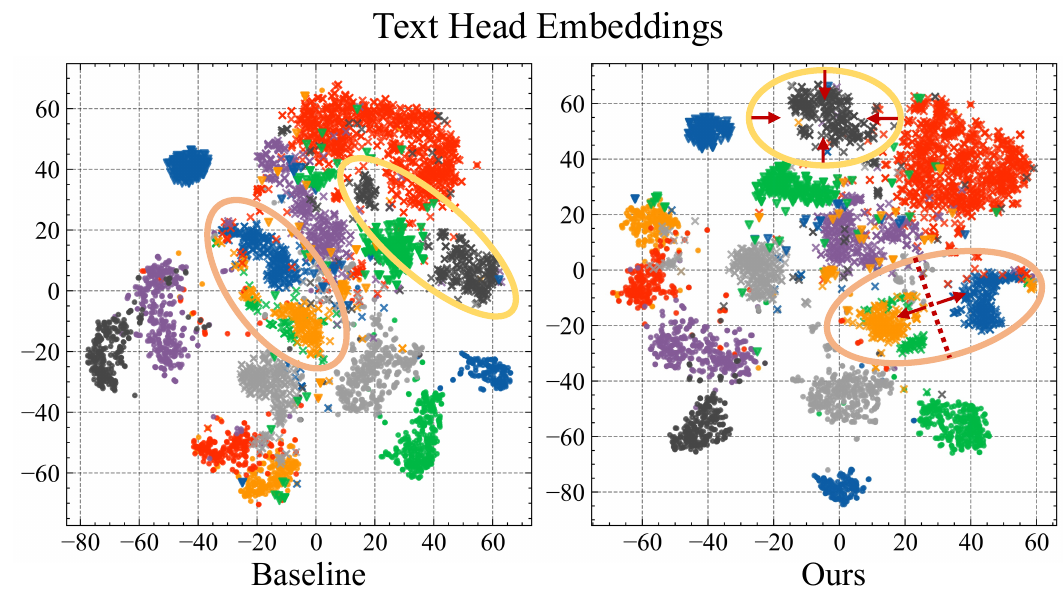}    
  \end{subfigure}
  \caption{The T-SNE visualization results of both RoI embeddings $\mathbf{f}_r$ in Equation~\ref{equation_roi_feature} and text embeddings $\mathbf{t}_r$ in Equation~\ref{equation_text_emb} between baseline and our proposed model for novel classes on OV-COCO benchmark. Best viewed with color. } 
  \label{figure_tsne}  
\end{figure*}   

\begin{figure*}[!tb]  
 \centering 
 \includegraphics[width=0.99\textwidth] {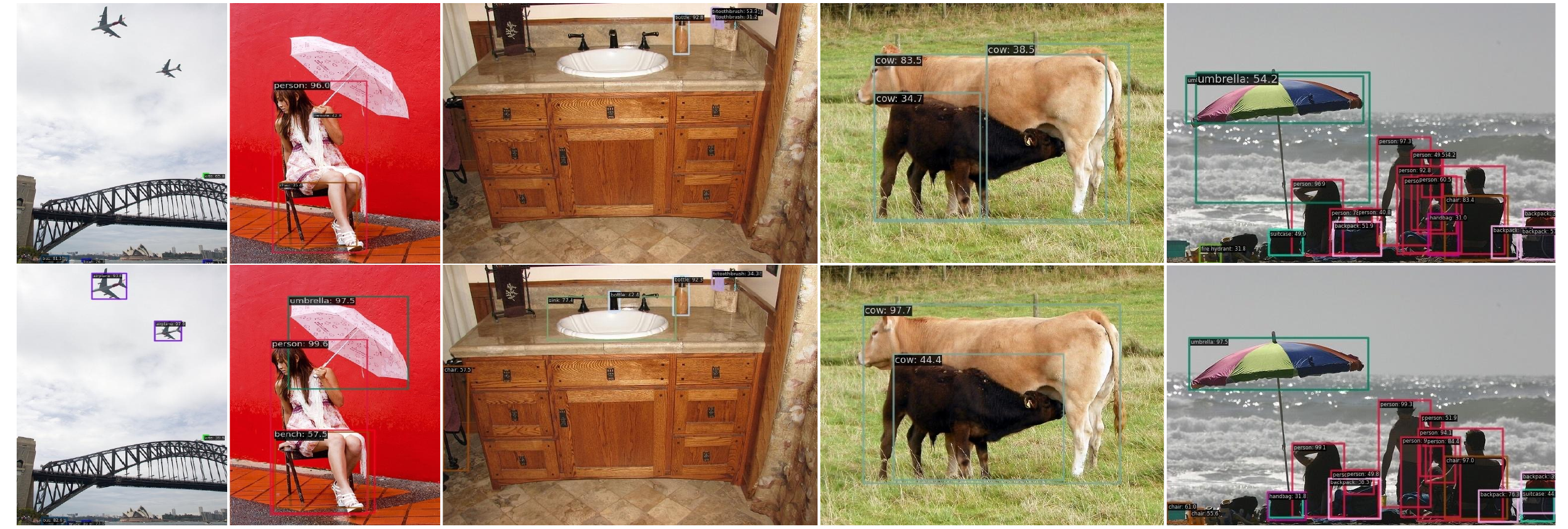}  
 \caption{The qualitative detection visualization results of baseline (Top) and our proposed model (Bottom) on OV-COCO benchmark, where airplane, umbrella, sink and cow are novel categories not seen during training. The left three columns represent a comparison of whether the detectors have missed any novel categories, while the right two columns represent a comparison of the detectors' accurate predictions and confidence of novel bounding boxes. }  
\label{figure_det_resultes_compare_vis}    
\end{figure*}   
 
\subsection{Further Exploration}  

In this subsection, we conduct an in-depth investigation of our model, including analysis of NRA attention weight, t-SNE visualization, detection results visualization.

\noindent\textbf{Attention weight visualization.} We visualize the attention weights within the NRA module to explore the interaction patterns between different regions. As shown in Figure~\ref{figure_attn_weight_vis}, a region patch is distinctly represented by six tokens ($l=6$) in the heatmap. Notably, the conspicuous responses in the principal diagonal elements elucidate the fundamental nature of self-region interaction, thereby facilitating the extraction of the region's inherent visual concepts. Then, it is evident that tokens within a region demonstrate connections, not only internally but also externally with other tokens. This underscores our model's adeptness in capturing relationships among regions,  and it can be further employed for distillation alignment to facilitate open-vocabulary detection.

\noindent\textbf{t-SNE visualization.} We employ t-SNE~\cite{van2008visualizing} to visually analyze
both RoI embeddings $\mathbf{f}_r$ in Equation~\ref{equation_roi_feature} and text embeddings $\mathbf{t}_r$ in Equation~\ref{equation_text_emb}. Figure~\ref{figure_tsne} illustrates comparative results between the baseline and our model for novel classes on OV-COCO, and distinct marker symbols and colors are utilized to symbolize different class samples. Overall, both baseline and our proposed model exhibit the capability to form clusters for novel categories. Moreover, as delineated by the ellipses in Figure~\ref{figure_tsne}, our NRAA model shows a more advanced capability in learning to discriminate features between different classes and promoting proximity among features of the same class, further proving its success.  

\noindent\textbf{Detection results visualization. } We present visualizations of open-vocabulary detection results in Figure~\ref{figure_det_resultes_compare_vis}. The left three columns represent a comparison of whether the detectors have missed any novel categories. It is evident that our model accurately identifies novel categories. For the right two columns, we similarly observe that the baseline model demonstrates recognition efficacy for novel categories. However, our model exhibits superiority in terms of the accuracy and confidence of novel bounding box predictions. 
  
\section{Discussion}

\noindent\textbf{Limitations and future work.} (\romannumeral1) Our model harnesses the pre-trained vision-language model, CLIP, and the performance of the detector depends on  this choice. The exploration of a more robust VLM stands as a prospective direction for future advancements. (\romannumeral2) Despite the seamless discard capability of the NRA module during the inference, without incurring additional computational overhead, it, akin to prior work, still requires the forward propagation of the text encoder. A promising direction for future research lies in the development of a more lightweight OVD detector leveraging our NRA mechanism. 

\section{Conclusion} 

In this paper, we introduce a novel open-vocabulary object detector, termed Neighboring Region Attention Alignment (NRAA). Specifically, our NRAA model firstly utilizes an attention mechanism to obtain interactive information among a set of regions. Subsequently, both the relation information and original visual concepts are fed into the vision-language models (VLMs) for distillation alignment and achieve open-vocabulary inference. Extensive experiments show that our model obtains superior performance on OVD benchmarks.

\backmatter

\bmhead{Acknowledgements}

This work was supported by the National Key Research and Development Plan under Grant 2021YFE0205700, Science and Technology Development Fund of Macau (0070$/$2020$/$AMJ, 0096$/$2023$/$RIA2), and Guangdong Provincial Key R$\&$D Programme: 2019B010148001.

\section*{Data Availability Statement} 

For this paper only publicly available datasets were used. The links of COCO and LVIS can be found in \url{https://cocodataset.org/#home} and \url{https://www.lvisdataset.org/}.

\section*{Declarations}

Conflict of interest. The authors have no conflict of interest to declare that are relevant to the content of this article.

\bibliography{sn-bibliography}

\end{document}